\documentclass[letterpaper, 10 pt, conference]{ieeeconf}  

\IEEEoverridecommandlockouts                              

\usepackage{amsmath} 
\usepackage{amssymb}  
\usepackage{tensor}
\usepackage{siunitx}

\usepackage{cite}
\usepackage{url}
\usepackage{siunitx}
\usepackage{color}
\usepackage{times}

\usepackage{epsfig}
\usepackage{svg}
\usepackage{balance}
\usepackage{graphicx}
\usepackage{amsfonts}
\usepackage{fancyhdr}
\usepackage{comment}
\usepackage{changepage}
\usepackage{bm}
\usepackage{calligra}
\usepackage{tabularx}
\usepackage{mathtools}
\usepackage{arydshln}
\usepackage{latexsym} 
\usepackage{float}
\usepackage{tabstackengine}
\usepackage{siunitx}
\usepackage{colortbl}
\usepackage{hhline}
\usepackage{bookmark}
\usepackage{dirtree}
\usepackage{subfigure}

\usepackage{booktabs} 
\usepackage{siunitx} 

\usepackage{tikz}
\usetikzlibrary{shapes.geometric, arrows}
\usepackage{graphicx}
\usepackage{algpseudocode}
\usepackage{algorithm}

\usepackage{booktabs}
\usepackage{xcolor}
\usepackage{multirow}

\usepackage{hyperref}

\title{\LARGE \bf
PALoc: Robust Prior-assisted Trajectory Generation for Benchmarking
}

\author{
Xiangcheng Hu, Jin Wu, Jianhao Jiao, Ruoyu Geng, Ming Liu
\thanks{X. Hu, J. Wu, J. Jiao, R. Geng and M. Liu are with Department of Electronic and Computer Engineering, Hong Kong University of Science and Technology, Hong Kong, China (E-mail: eelium@ust.hk)  
}
}

\begin{document}

\maketitle

\begin{abstract}
  Evaluating simultaneous localization and mapping (SLAM) algorithms necessitates high-precision and dense ground truth (GT) trajectories.
  But obtaining desirable GT trajectories is sometimes challenging without GT tracking sensors.
  As an alternative, in this paper, we propose a novel prior-assisted SLAM system to generate a full six-degree-of-freedom ($6$-DOF) trajectory at around $10$Hz for benchmarking, under the framework of the factor graph.
  Our degeneracy-aware map factor utilizes a prior point cloud map and LiDAR frame for point-to-plane optimization, simultaneously detecting degeneration cases to reduce drift and enhancing the consistency of pose estimation.
  Our system is seamlessly integrated with cutting-edge odometry via a loosely coupled scheme to generate high-rate and precise trajectories.
  Moreover, we propose a norm-constrained gravity factor for stationary cases, optimizing both pose and gravity to boost performance.
  Extensive evaluations demonstrate our algorithm's superiority over existing SLAM or map-based methods in diverse scenarios, in terms of precision, smoothness, and robustness.
  Our approach substantially advances reliable and accurate SLAM evaluation methods, fostering progress in robotics research.
\end{abstract}

\section{Introduction}

SLAM algorithm evaluation motivates the need for reliable trajectory generation.
However, acquiring dense, smooth, and accurate 6-DOF trajectory poses still remains challenging.
The first category of trajectory generation methods is based on the tracking of markers such as the motion capture system (MOCAP) \cite{mueggler2017event} and global navigation satellite system (GNSS) \cite{Geiger2012}, but they are constrained in laboratories and outdoor roads with the rich satellite signal.
Although the laser tracking devices \cite{nguyen2022ntu} are not limited in specific environments, they require the prism to be always observable during a sequence.

The other category regards the prior map-based localization methods \cite{Jiao2022Mar, Zhang2022Aug}.
They support the pose generation in a wide range of scenarios if a pre-collected point cloud map or building information modeling (BIM) is available.
Their overall approach is to use range sensors, i.e., LiDARs and RGB-D cameras, to address the frame-to-map alignment problem using registration algorithms.
However, most of them neglect the noise nature of range sensors and only consider spatial information, and cannot handle issues such as sensor degeneration, intense movements, and poor local smoothness in nonideal environments.
To address these challenges, we propose a flexible prior-assisted localization approach with a global factor graph. Building upon our prior work \cite{Jiao2022Mar}, Our contributions are as follows:
\begin{itemize}
  \item We develop a prior-assisted localization system that combines a prior map with local sensor measurements, facilitating the generation of 6-DOF dense poses without the need for specialized GT tracking sensors.
  \item We propose a degeneracy-aware map factor to address common degeneration cases by considering the coupling of eigenvalues and eigenvectors along with translation constraint strength for continuous pose estimation.
  \item We introduce a norm-constrained gravity factor specifically tailored for Zero Velocity Update (ZUPT) scenarios, optimizing both pose and gravity simultaneously.
\end{itemize}

\begin{figure}[t]
  \centering
  \subfigure[ ]{
    \label{fig:sensor_suit}
    \includegraphics[width=.378\linewidth]{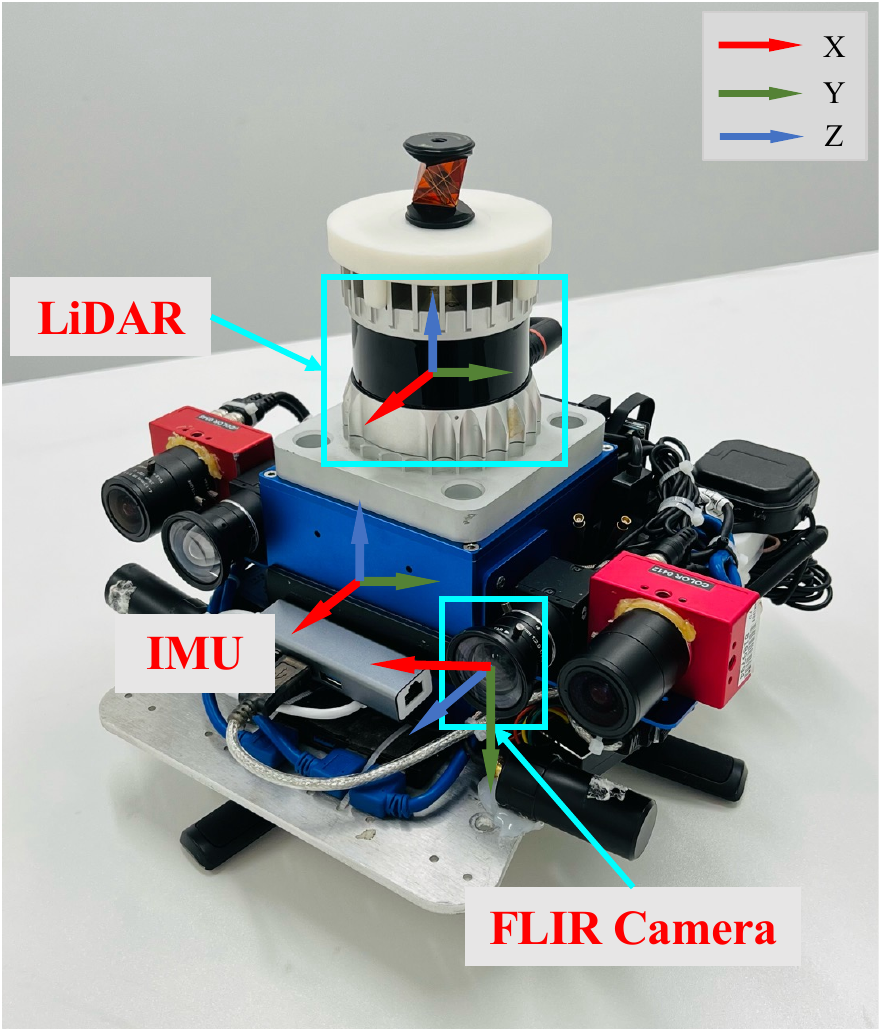}
  }
  \subfigure[ ]{
    \label{fig:sensor_kit_mcr}
    \includegraphics[width=.432\linewidth]{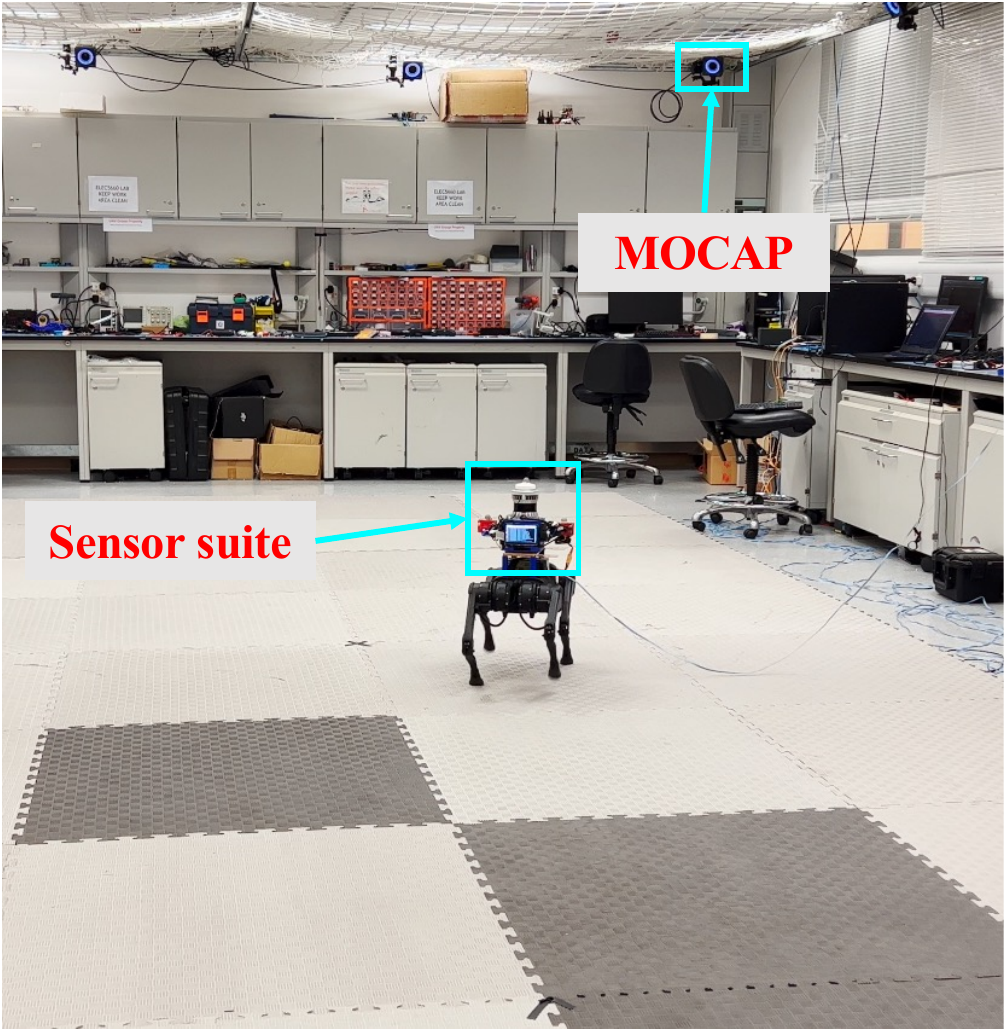}
  }
  \subfigure[ ]{
    \label{fig:task_description2}
    \includegraphics[width=.95\linewidth]{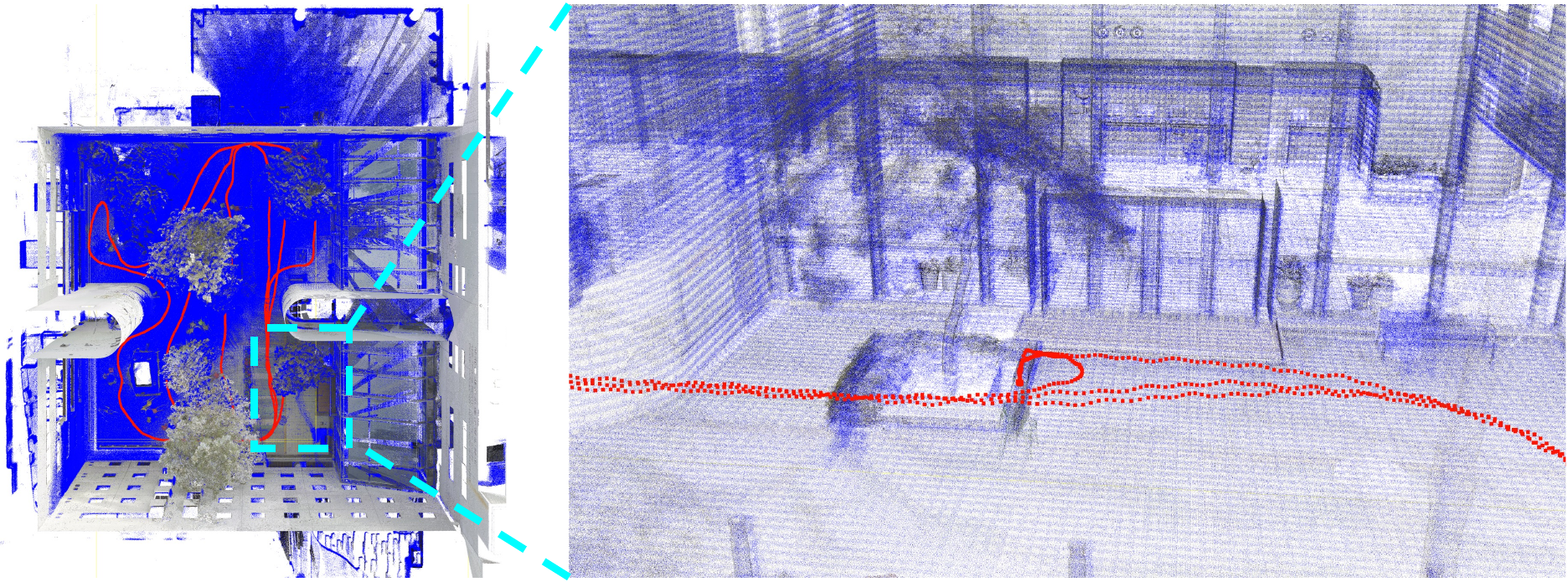}
  }
  \caption{(a) Sensors configuration with corresponding coordinate frames. (b) Quadruped robot equipped with sensor suite in Motion Capture Room (MCR). (c) Prior RGB point cloud map with the estimated trajectory (red points) and map (blue point cloud) by PALoc.}
  \label{fig:sensor_kit}
  \vspace{-2em}
\end{figure}

\section{Methodology} \label{sec:problem_statement}

\subsection{Notations and Definitions}\label{sec:notations}
We formulate the trajectory generation problem in our system, which includes a LiDAR, an IMU, and a prior  map. The body frame, denoted as $\bm{()^b}$, is defined by the IMU, and the global frame, represented by $\bm{()^w}$, is determined by the prior map. The robot's pose at time $k$ is expressed as $\bm{p}_k = \bm(\bm{t}_k, \bm{R}_k)$, with $\bm{t}_k$ indicating the position and $\bm{R}_k$ representing the orientation as a rotation matrix. The robot's velocity at time $k$ is symbolized by $\bm{v}_{k}$, and accelerometer and gyroscope biases at time $k$ are represented by $\bm{b}_{a,k}$ and $\bm{b}_{\omega,k}$. The complete state vector is defined: $\bm{\mathcal{X}}= [\bm{R}, \bm{t}, \bm{v}, \bm{b}_{a}, \bm{b}_{\omega}, \bm{g}]$.
\begin{figure}[t]
  \centering
  \includegraphics[width=0.45\textwidth]{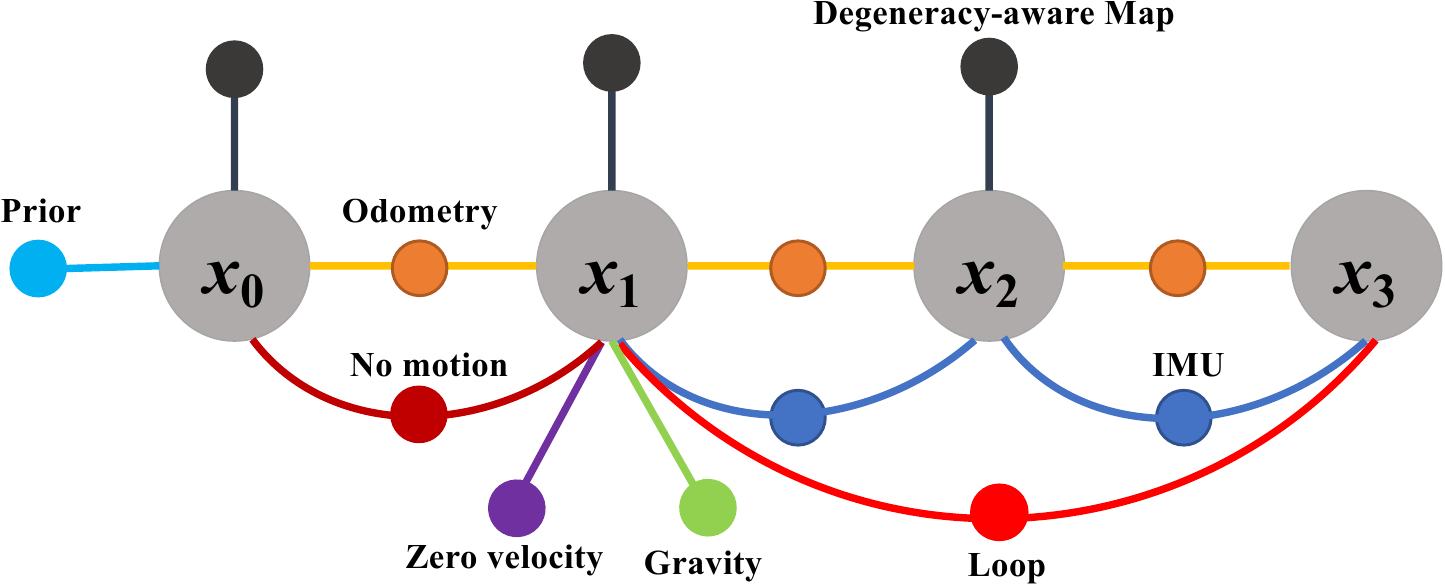}
  \caption{Factor graph of PALoc. The diagram illustrates the graph structure with diverse nodes and factors. The large gray circle signifies the system state $\bm{x}_i$, and the smaller colored circles denote distinct factors.}
  \label{fig:factor_graph}
  \vspace{-0.5cm}
\end{figure}

\subsection{Factor Graph Formulation}\label{sec:formulation}
We abstract the pose estimation problem of the system with a factor graph, as shown in Fig. \ref{fig:factor_graph}.
The graph: $\mathcal{G} = (\mathcal{X}, \mathcal{F}, \mathcal{E})$, where $\mathcal{X}$ is the set of state variables, $\mathcal{F}$ denotes constraints between variables, and $\mathcal{E}$ signifies edges connecting factors and variables.
The factor graph includes several factors: LiDAR Odometry Factor (LO), IMU Factor (IM), Zero Velocity Factor (ZV), No Motion Pose Factor (NM), Norm-Constrained Gravity Factor (NG), and Degenracy-aware Map Factor (DM).

\subsubsection{Norm-Constrained Gravity Factor (NG)}
The norm-constrained gravity factor enforces constraints on gravity's magnitude and direction under ZUPT conditions.
We minimize the gravity direction error while ensuring $||\bm{g}||=1$.
We transform the measured acceleration vector, $\bm{a}_{\text {m}}^b$, into the world frame as $\bm{a}^w$ with rotation $\bm{R}$:
$\bm{a}_{\text {m}}^b=-\bm{g}$, $\bm{a}^w=\bm{R} \bm{a}_m^b$.
We define the direction error $\bm{e}_{\text{dir}}$ and the magnitude error $\bm{e}_{\text{mag}}$ (the $z$-axis of the equipped IMU is up) respectively:
$\bm{e}_{\text{dir}}=\bm{a}^w/\left\|\bm{a}^w\right\|+\bm{g}, \bm{e}_{\text{mag}}=\|\bm{g}\|-1$.

\subsubsection{Degenracy-Aware Map Factor (DM)}
DM constraints the pose of a robot to align with the prior map.
Point-to-plane matching is used in the proposed approach, and a DM is designed to detect state degeneration \cite{Tuna2022Nov,jiao2021Greedy} and selectively added to the factor graph.
To circumvent the degeneration issue, we introduce a two-stage degeneration detection approach. In the first stage, the spectrum property $D_{e}$ \cite{jiao2021Greedy} of the Hessian matrix $\bm{H}=\bm{J^TJ}$ with respect to the pose is computed and compared with threshold:
\begin{equation}
  D_e=\sum_{i=1}^6 \frac{1}{\lambda_i}\left(1-\frac{\left|e_i^T v_i\right|}{\left|e_i\right|_2\left|v_i\right|_2}\right)^2,
\end{equation}
where $\lambda_i$ corresponds to the $i$-th eigenvalue, while $\mathbf{e}_i$ and $\mathbf{v}i$ represent the $i$-th eigenvectors of the measurement and reference point clouds, respectively. Let's focus on each corresponding point pair. The Jacobian matrix is analyzed to examine constraint impacts along translation directions and divide points into groups \cite{Tuna2022Nov}. The dimension with the least constraints is determined, and ratio factors, $s$, are defined as $s_i = N_{i}/N_{min}$, where $i\in \{x, y, z\}$. With a set threshold, $s_{thres}$, constraint ratios ($s_i$) are compared. If $s_i < s_{thres}$, the corresponding dimension is classified as degeneracy.

\begin{figure}
  \centering
  \includegraphics[width=0.47\textwidth]{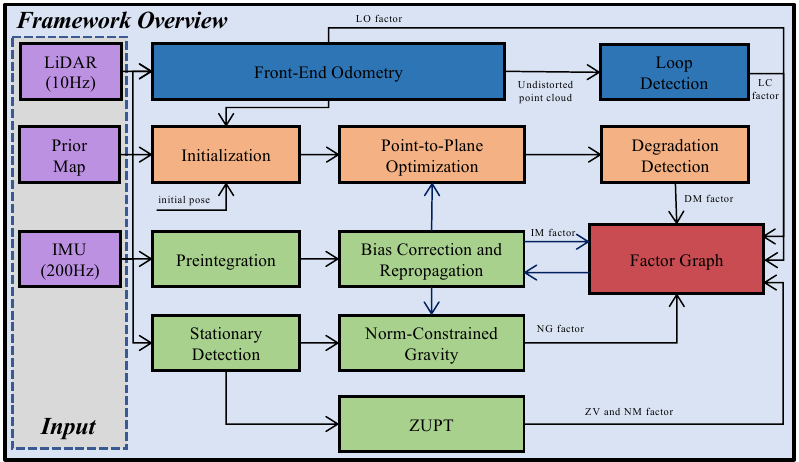}
  \caption{The general diagram of PALoc.}
  \label{fig:pipeline}
  \vspace{-0.5cm}
\end{figure}

\begin{figure*}[t]
  \centering
  \subfigure[]{
    \label{fig:traj_corridor_3d}
    \includegraphics[width=.4\linewidth]{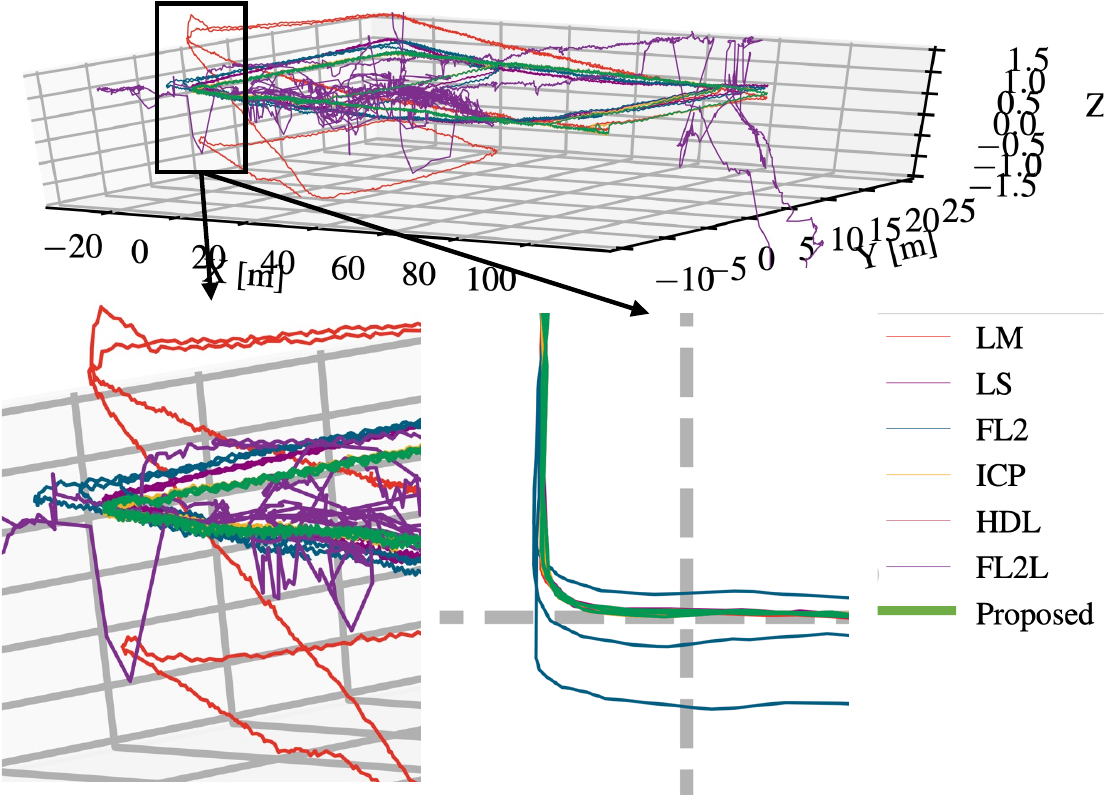}
  }
  \subfigure[]{
    \label{fig:traj_MCR_slow}
    \includegraphics[width=.4\linewidth]{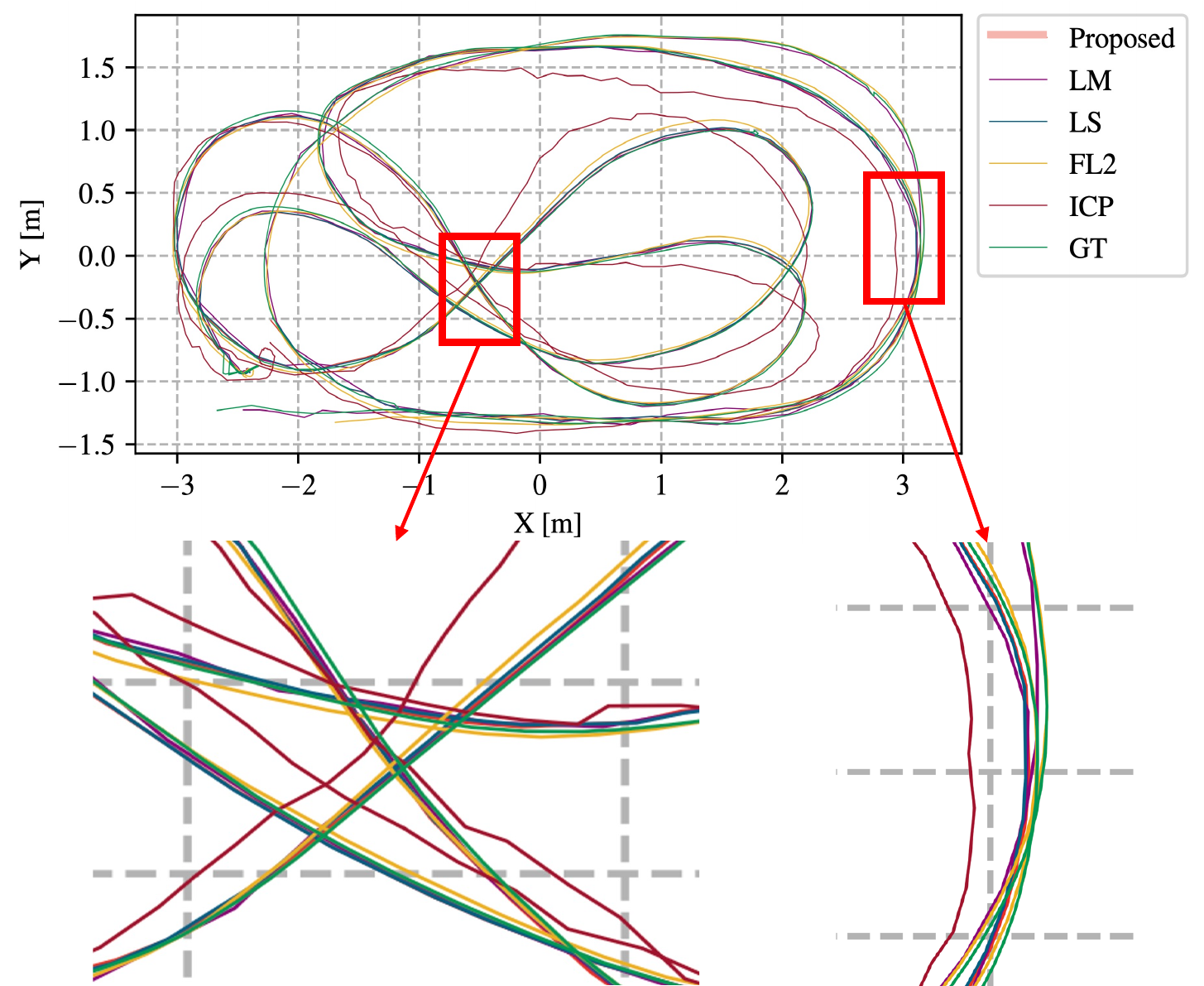}
  }
  \caption{Trajectories alignment with SOTA algorithms. (a) X-Z, X-Y views of corridor\_day. (b) X-Y view of two distinct areas of MCR\_slow.}
  \label{fig:corridor_aligened_traj}
  \vspace{-0.5cm}
\end{figure*}

\begin{figure*}[t]
  \setlength{\subfigcapskip}{-0.1cm}
  \setlength{\subfigbottomskip}{-0.1cm}
  \centering
  \subfigure[]{
    \label{fig:corridor}
    \includegraphics[width=0.24\linewidth]{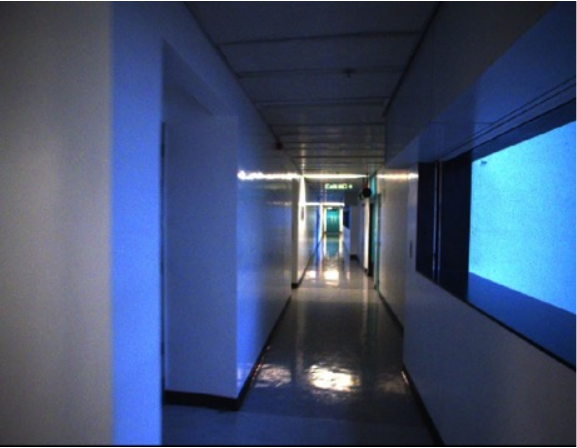}
  }
  \subfigure[Intersection with balanced constraints]{
    \label{fig:corridor_normal_intersections}
    \includegraphics[width=0.33\linewidth]{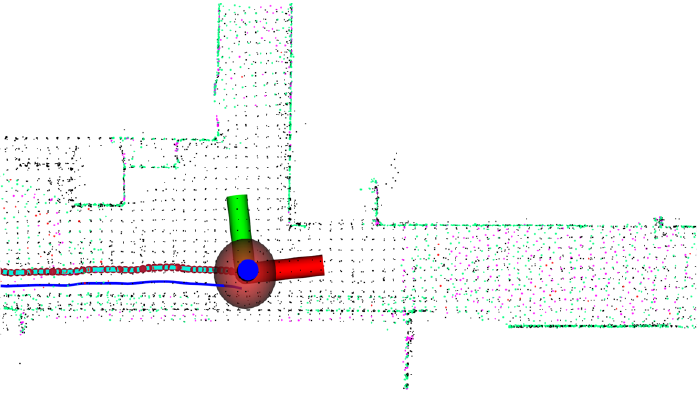}
  }
  \subfigure[ Severe z-axis drift in U-turn intersection]{
    \label{fig:corridor_z_axis}
    \includegraphics[width=0.33\linewidth]{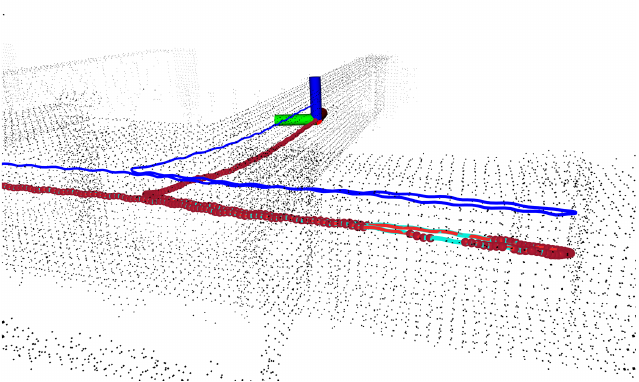}
  }
  \quad
  \subfigure[]{
    \label{fig:corridor_degenerated}
    \includegraphics[width=0.24\linewidth]{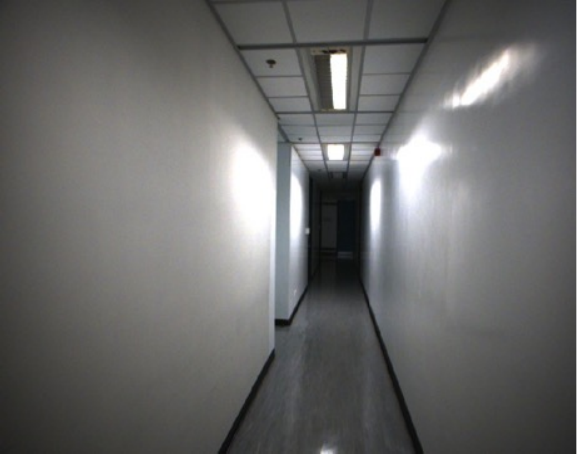}
  }
  \subfigure[Narrow corridor with degenerated $x$-dimension]{
    \label{fig:corridor_normal}
    \includegraphics[width=0.33\linewidth]{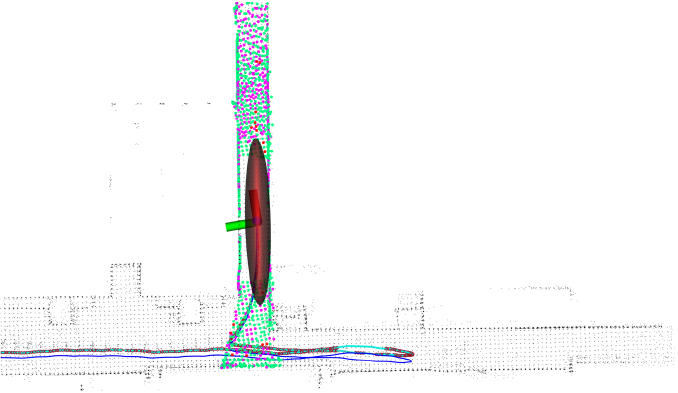}
  }
  \subfigure[Point clouds constraint strength classification]{
    \label{fig:corrdiro_dengenratin_detailed}
    \includegraphics[width=0.33\linewidth]{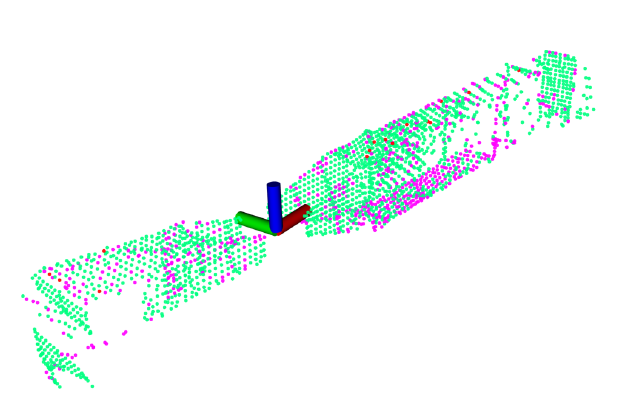}
  }
  \caption{Degeneration analysis on corridor\_day. (a) Scene image of the structure-rich corridor. (d) Scene image of the narrow and structureless long corridor.
    (b),(c),(e),(f) The black point cloud represents the prior map, and the red sphere with coordinate axes represents the relative constraint strength in the $XYZ$ dimensions, but unrelated to the overall size of the ellipsoid.
    The flatter the ellipsoid, the more severe the degeneration in a specific dimension. The blue and light blue trajectories and the red points on the trajectories represent the FL2 trajectory, our algorithm trajectory, and the pose with map factor constraints. Our algorithm easily eliminates $z$-axis drift error while ensuring robustness in a U-turn intersection (c). The point clouds of different colors in (f) indicate the corresponding number of constraints in $XYZ$ dimensions.}
  \label{fig:degeneration_case}
  \vspace{-0.2cm}
\end{figure*}

\subsection{PALoc}
\label{sec:system_overview}
We develop our system under two assumptions:
\begin{enumerate}
  \item Sensors are time-synchronized in hardware, ensuring precise data alignment.
  \item We focus on Pose-SLAM, which means that optimal poses result in the best-built map.
\end{enumerate}

Fig. \ref{fig:pipeline} illustrates the pipeline of the proposed PALoc.
The process starts with a front-end odometry that provides an undistorted point cloud and global pose of each frame while generating an initial pose to align with the prior map, effectively initializing the entire system.
The LO and DM introduce local and global constraints, respectively.
ZUPT and gravity constraints are employed to further improve the localization accuracy under stationary conditions.

\begin{table*}[t]
\centering
\caption{comparison of \textbf{ate} (\SI{}{\cm}) and \textbf{rpe} (\SI{}{\cm}) for different algorithms on different platform sequences.}
\renewcommand\arraystretch{1.1}
\renewcommand\tabcolsep{2pt}
\small
\begin{tabular}{l c c c c c|c|c|c|c|c|c|c|c|c|c|c}
\toprule[0.03cm]
\textbf{Sequence}
 & \multicolumn{2}{c|}{\textbf{LM}} & \multicolumn{2}{c|}{\textbf{LS}} & \multicolumn{2}{c|}{\textbf{FL2}} & \multicolumn{2}{c|}{\textbf{ICP}} & \multicolumn{2}{c|}{\textbf{HDL}} & \multicolumn{2}{c|}{\textbf{FL2L}} & \multicolumn{2}{c|}{\textbf{Proposed-LS}} & \multicolumn{2}{c}{\textbf{Proposed-FL2}} \\
\cline{2-17}
& \textbf{ATE}$\downarrow$ & \textbf{RPE}$\downarrow$ & \textbf{ATE}$\downarrow$ & \textbf{RPE}$\downarrow$ & \textbf{ATE}$\downarrow$ & \textbf{RPE}$\downarrow$ & \textbf{ATE}$\downarrow$ & \textbf{RPE}$\downarrow$ & \textbf{ATE}$\downarrow$ & \textbf{RPE}$\downarrow$ & \textbf{ATE}$\downarrow$ & \textbf{RPE}$\downarrow$ & \textbf{ATE}$\downarrow$ & \textbf{RPE}$\downarrow$ & \textbf{ATE}$\downarrow$ & \textbf{RPE}$\downarrow$ \\
\midrule[0.03cm]
MCR\_slow & \textbf{3.74} & 4.76 & \underline{6.19} & 4.21 & 13.70 & \underline{2.79} & 8.42 & 4.90 & $\times$ & $\times$ & $\times$ & $\times$  & \underline{5.70} & 3.89 & 7.36 & \textbf{2.49} \\
MCR\_normal & \underline{8.03} & 5.13 &  8.16 & \underline{4.94} & $\times$ & $\times$ & $\times$ & $\times$ & $\times$ & $\times$ & $\times$ & $\times$ & \textbf{7.71} &  4.95 & 10.12 & \textbf{3.78}  \\
\midrule[0.03cm]
MCR\_slow\_00 & \underline{2.15} & 1.21 & 2.34 & 0.90 & 3.81 & \textbf{0.52} & 3.89 & 1.25 & $\times$ & $\times$ & 4.61 & 3.70  &  \textbf{ 2.00} & 0.90 & 3.73 & \underline{0.54}\\

MCR\_slow\_01 & \textbf{2.62}  & 7.04 & \underline{2.75} & 1.58 & 3.68  & \textbf{0.59} & 3.11  & 1.60 & $\times$  & $\times$ & 4.97 & 5.91  & 2.93 & 1.51 & 2.81  & \underline{0.70}\\
MCR\_normal\_00 & $\times$  & $\times$ & \underline{4.02}  & 2.48 & 9.80  & \textbf{1.16} & $\times$  & $\times$ & $\times$  & $\times$  & 9.55  & 10.46 & \textbf{3.96} & 2.49 & 5.03  & \underline{1.19} \\
MCR\_normal\_01 & 12.73  & 6.9 & \textbf{3.53}  & \textbf{0.89} & 5.99 & 1.09 & 11.81  & 5.05 & $\times$  & $\times$  &  10.43  & 11.02  & \underline{3.58} & \underline{0.93} & 5.24  & 1.06\\
\bottomrule[0.03cm]
  \multicolumn{9}{l}{
    $\times$: algorithms fail. \textbf{bold}: best results. \underline{underlined}: second-best results.
  }\\
\end{tabular}
\label{tab:traj_ape_rpe}
\vspace{-2em}
\end{table*}

\begin{table*}[t]
\centering
\caption{evaluation of map accuracy in terms of \textbf{acc} [\SI{}{\cm}] and \textbf{com} [\%] of the estimated point cloud map within \SI{20}{\cm} threshold.}
\renewcommand\arraystretch{1.1}
\renewcommand\tabcolsep{2pt}
\small
\begin{tabular}{l||c|c||c|c||c|c||c|c||c|c||c|c||c|c}

\toprule[0.03cm]
\textbf{Sequence} &
\multicolumn{2}{c||}{\textbf{LM}} &
\multicolumn{2}{c||}{\textbf{LS}} &
\multicolumn{2}{c||}{\textbf{FL2}} &
\multicolumn{2}{c||}{\textbf{ICP}} &
\multicolumn{2}{c||}{\textbf{HDL}} &
\multicolumn{2}{c||}{\textbf{FL2L}} &
\multicolumn{2}{c}{\textbf{Proposed-FL2}} \\
\cline{2-15}
 & \textbf{ACC}$\downarrow$ & \textbf{COM}$\uparrow$ & \textbf{ACC}$\downarrow$ & \textbf{COM}$\uparrow$ & \textbf{ACC}$\downarrow$ & \textbf{COM}$\uparrow$ & \textbf{ACC}$\downarrow$ & \textbf{COM}$\uparrow$ & \textbf{ACC}$\downarrow$ & \textbf{COM}$\uparrow$ & 
\textbf{ACC}$\downarrow$ & \textbf{COM}$\uparrow$ & \textbf{ACC}$\downarrow$ & \textbf{COM}$\uparrow$ \\
\midrule[0.03cm]
garden\_day & 4.14 & 93.52 & 3.94 & 95.46 & 5.98 & \underline{95.68}  & 3.64 & 94.79 & 6.06 & 95.61 & \underline{3.50} & 95.03 & \textbf{3.48} & \textbf{95.70} \\
garden\_night & 4.36 & 94.67 & 3.92 & 96.16 & 5.91 & 96.36 & \underline{3.23} & \textbf{96.60} & 6.12 & 96.22 & 3.52 & 95.44 & \textbf{3.19} & \underline{96.59} \\
canteen\_day & 5.65 & 77.01 & 5.48 & 78.63 & 6.32 & 81.57 & \underline{4.86} & \underline{82.15} & 6.59 & 81.30 & 5.59 & 80.23 & \textbf{4.71} & \textbf{82.16}  \\

canteen\_night & 5.60 & 76.08 & 5.29 & 79.97 & 6.77 & 81.27  & \underline{5.07} & \underline{82.39} & 6.93 & 80.97  &  5.56 & 81.14 & \textbf{4.76} & \textbf{82.54}\\

corridor\_day & 7.40 & 68.58 & 6.24 & 76.90 & 7.28 & 75.12  & $\times$ & $\times$ & \underline{5.04} & \underline{85.73} & $\times$  & $\times$ &\textbf{3.99} & \textbf{94.37}\\ 
escalator\_day & 5.40 & 89.58 & 8.85 & 52.61 & 6.92 & 83.84  & $\times$ & $\times$ & 7.66 & 90.12 & \underline{4.29} & \underline{93.23} & \textbf{3.88} & \textbf{93.26} \\
building\_day & 10.11 & 27.52 & 7.65 & 71.94 & 6.68 & 79.36 & 6.72  & 86.71  & 7.11 &  90.12  & \underline{4.19} & \underline{91.24} &  \textbf{4.14} & \textbf{93.35} \\
MCR\_slow & 7.66 & 49.66 & \underline{4.08} & 91.90 & 6.19 & 87.58 & \textbf{3.96} & \textbf{94.02}  & $\times$& $\times$  & $\times$ & $\times$ & 4.63 & \underline{93.75} \\
MCR\_normal & 4.28 & 89.81 & \underline{3.85} & \underline{91.61}  & $\times$ & $\times$   & $\times$ &  $\times$  & $\times$ & $\times$ & $\times$ & $\times$ & \textbf{3.71} & \textbf{91.90}\\

  \bottomrule[0.03cm]
 \multicolumn{9}{l}{
    $\times$: algorithms fail. \textbf{bold}: best results. \underline{underlined}: second-best results.
  }\\
  \end{tabular}
\label{tab:map_remse}
\vspace{-1.5em}
\end{table*}

\section{EXPERIMENTAL RESULTS}\label{sec:experimental_results}

\subsection{Experimental Setup}
\subsubsection{Experimental Setup}
Our experiments utilized a handheld sensor suite, which included an Ouster-128 OS1 LiDAR with measurement noise of \SI{3}{\cm}, a STIM300 IMU, and two FLIR RGB cameras (see Fig. \ref{fig:sensor_kit}). Prior maps were obtained by a Leica BLK360 laser scanner. For the experiments, we employed a high-performance desktop computer featuring an Intel i7 processor, 96 GB of DDR4 RAM, and 1 TB SSD storage.  We compared our method with several renowned SOTA LIO as well as map-based localization algorithms, including FAST-LIO2 (FL2)\footnote{\scriptsize\url{https://github.com/hku-mars/FAST_LIO}}, LIO-SAM (LS)\footnote{\scriptsize\url{https://github.com/JokerJohn/LIO_SAM_6AXIS}}, LIO-Mapping (LM)\footnote{\scriptsize\url{https://github.com/hyye/lio-mapping}}, HDL-Localization (HDL)\footnote{\scriptsize\url{https://github.com/koide3/hdl_localization}}, and ICP-Localization (ICP), FASTLIO-Localization (FL2L)\footnote{\scriptsize\url{https://github.com/HViktorTsoi/FAST_LIO_LOCALIZATION}}.
We employed FL2 and LS as the front-end odometry for experiments, named \textbf {Proposed-FL2} and \textbf{Proposed-LS}, respectively. We test algorithms with the FusionPortable dataset \cite{Jiao2022Mar}.

\subsubsection{Evaluation Metrics}
We employed widely-used metrics such as Absolute Trajectory Error (ATE) and Relative Pose Error (RPE) for trajectory evaluation. For map evaluation, we utilized metrics such as completeness and accuracy \cite{Aanaes2016Nov}. Completeness (COM) measures the proportion of true points successfully matched with the GT map, while accuracy (ACC) computed the average Euclidean distance between estimated points and points from the GT map.

\subsection{Trajectory Evaluation}
Table \ref{tab:traj_ape_rpe} shows the ATE and RPE evaluation results of our Proposed-LS and Proposed-FL2 algorithms on handheld and quadruped platforms in the MCR room, compared to SOTA algorithms. Our algorithm achieves a perfect balance of accuracy and smoothness compared to the original LS and FL2 algorithms. In comparison to other SLAM, prior-assisted SLAM, and Map-based methods, our algorithm successfully generates higher-precision trajectories regardless of the motion pattern, further demonstrating the accuracy and robustness of our algorithm. Fig. \ref{fig:traj_MCR_slow} illustrates the trajectory alignment performance of our algorithm on MCR\_slow.

\begin{figure}[t]
  \setlength{\subfigcapskip}{-0.05cm}
  \setlength{\subfigbottomskip}{-0.05cm}
  \subfigure[Top and X-Z view of escalator\_day with ceiling removal]{
    \label{fig:distance_error_map_escaltor}
    \includegraphics[width=.95\linewidth,height=0.15\textwidth]{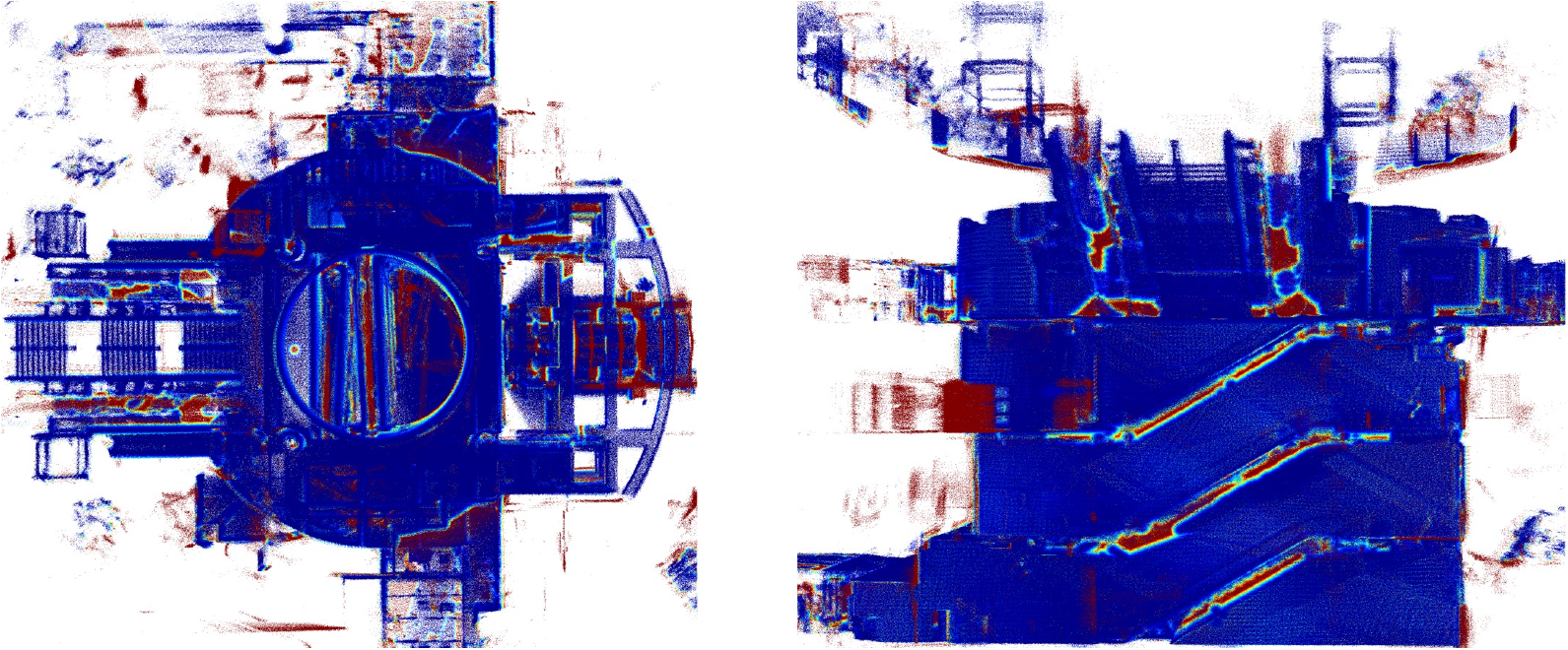}
  }
  \subfigure[corridor\_day]{
    \label{fig:distance_error_map_corridor}
    \includegraphics[width=.95\linewidth,height=0.16\textwidth]{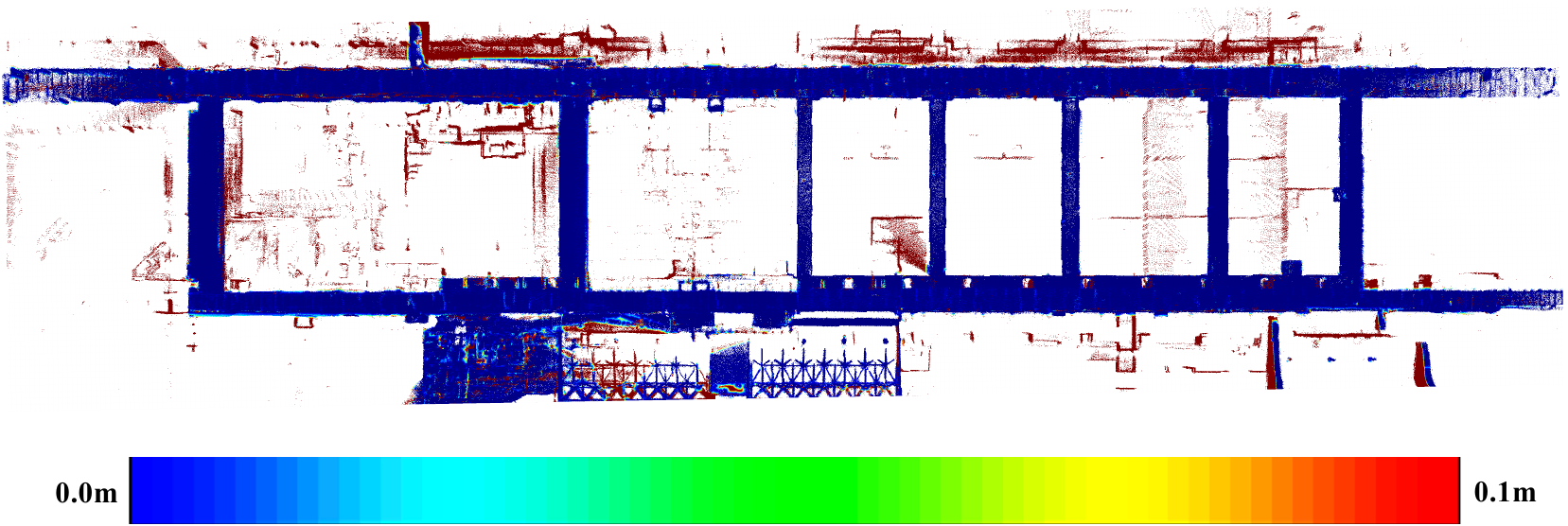}
  }
  \caption{Distance error map of different sequences.}
  \label{fig:distance_error_map_all}
  \vspace{-0.5cm}
\end{figure}

\subsection{Map Evaluation} \label{sec:map_evaluation}

Tables \ref{tab:map_remse} illustrate its robustness and accuracy across diverse campus scenes and platforms.
Our algorithm operates smoothly and maintains minimal map accuracy in long-distance degeneration scenarios, such as corridor\_day and escalator\_day, with significant $z$-axis height variation. Fig. \ref{fig:degeneration_case} presents the degeneration analysis in corridor\_day data, particularly in the narrow corridor (Fig. \ref{fig:corridor_normal}), where degeneration is detected, and high-precision pose estimation is sustained.
Fig. \ref{fig:corrdiro_dengenratin_detailed} examines degeneration caused by minimal laser points constraining the $x$- and $z$-directions, resulting in z-axis accumulation error in the narrow corridor.
Fig. \ref{fig:corridor_z_axis} demonstrates our algorithm's ability to effortlessly eliminate z-axis accumulation error compared to the original odometry.
Fig. \ref{fig:distance_error_map_all} displays error area distribution between the map estimated by our algorithm on the several sequences and the prior map, with a threshold of less than $0.1m$. Most areas estimated by our algorithm maintain an accuracy of nearly $3cm$, validating our trajectory accuracy and effectiveness.



\subsection{Run-time Evaluation}
We evaluate the computation time of the proposed system on corridor\_day. The FL2 module requires $38ms$ per frame. The DM module needs $141.8ms$ per frame, while the ZUPT-related Factor module necessitates $0.1ms$ per frame.

\section{Conclusion}\label{sec:conclusion}
In this paper, we introduced a prior-assisted localization system for generating dense trajectories to evaluate SLAM algorithms. Our system combines prior map constraints, LiDAR-based odometry, a universal factor graph, a DM, and an NG to enhance pose estimation robustness and accuracy. Future work involves improving efficiency for larger scenes and examining system observability and pose uncertainty. Our approach contributes to SLAM algorithm evaluation and advances the field of robotics and autonomous systems.




\bibliographystyle{IEEEtran}
\bibliography{IEEEexample.bib}

\addtolength{\textheight}{-12cm}   

\end{document}